# When Agents Fail to Act: A Diagnostic Framework for Tool Invocation Reliability in Multi-Agent LLM Systems

Donghao Huang[1,2], Gauri Malwe[3], Zhaoxia Wang[1]
[1]School of Computing and Information Systems, Singapore Management University, Singapore
[2]Research and Development, Mastercard, Arlington, VA, USA
[3]Research and Development, Mastercard, Pune, India
dh.huang.2023@engd.smu.edu.sg, gauri.malwe@mastercard.com, zxwang@smu.edu.sg

*Abstract*—Multi-agent systems powered by large language models (LLMs) are transforming enterprise automation, yet systematic evaluation methodologies for assessing tool-use reliability remain underdeveloped. We introduce a comprehensive diagnostic framework that leverages big data analytics to evaluate procedural reliability in intelligent agent systems, addressing critical needs for SME-centric deployment in privacy-sensitive environments. Our approach features a 12-category error taxonomy capturing failure modes across tool initialization, parameter handling, execution, and result interpretation. Through systematic evaluation of 1,980 deterministic test instances spanning both open-weight models (Qwen2.5 series, Functionary) and proprietary alternatives (GPT-4, Claude 3.5/3.7) across diverse edge hardware configurations, we identify actionable reliability thresholds for production deployment. Our analysis reveals that procedural reliability—particularly tool initialization failures—constitutes the primary bottleneck for smaller models, while qwen2.5:32b achieves flawless performance matching GPT-4.1. The framework demonstrates that mid-sized models (qwen2.5:14b) offer practical accuracy-efficiency trade-offs on commodity hardware (96.6% success rate, 7.3s latency), enabling cost-effective intelligent agent deployment for resource-constrained organizations. This work establishes foundational infrastructure for systematic reliability evaluation of tool-augmented multi-agent AI systems.

*Keywords—intelligent agents, multi-agent systems, large language models, tool-use reliability, error taxonomy*

## I. INTRODUCTION

Multi-agent systems leveraging large language models (LLMs) for autonomous task execution represent a significant advancement in artificial intelligence [1], [2], yet their production deployment remains constrained by fundamental challenges in procedural reliability and systematic evaluation. While existing research demonstrates tool-augmented LLM capabilities across various domains [3]-[5], the field critically lacks comprehensive diagnostic frameworks for characterizing failure modes in multi-agent coordination scenarios where tools must be reliably invoked, parameterized, and validated.

Current evaluation approaches predominantly report aggregate task success rates [6], [7], obscuring underlying failure causes and limiting opportunities for targeted improvements. Recent work by Kokane et al. [8] demonstrates the value of granular error taxonomies for single-agent tool use, yet these frameworks remain insufficient for multi-agent contexts where failures cascade through coordination protocols. Chawla et al. [9] identify tool use as a fundamental pillar of agentic AI, yet systematic reliability evaluation methodologies for multi-agent systems remain critically underdeveloped.

We address this gap by establishing a diagnostic framework that moves beyond aggregate metrics to provide actionable insights into specific failure patterns. Recognizing that organizations—particularly small and medium-sized enterprises (SMEs)—face diverse deployment constraints including hardware limitations, privacy requirements, and cost sensitivity [10], we also propose a tiered evaluation that maps reliability–cost trade-offs across model scales and hardware configurations.

Our contributions are:

**1) Systematic Diagnostic Framework:** A novel 12-category error taxonomy spanning tool initialization, parameter handling, execution, and result interpretation, extending single-agent diagnostic capabilities [8] to multi-agent coordination scenarios.

**2) Reproducible Evaluation Infrastructure:** Standardized protocols comparing open-weight models (qwen2.5:3b through 72B, Functionary) against closed-source alternatives (GPT-4, Claude 3.5/3.7) across diverse edge hardware (NVIDIA RTX A6000, RTX 4090, Apple M3 Max) using 4-bit quantization for realistic deployment conditions.

**3) Reliability Threshold Identification:** Quantitative analysis across 1,980 deterministic test instances identifying critical capacity thresholds—qwen2.5:32b achieves flawless reliability matching GPT-4.1, while qwen2.5:14b represents the minimum viable production configuration (96.6–97.4% success).

**4) Deployment Guidance and Open Resources:** Hardware-performance characterization demonstrating 8.2× latency variation across platforms, establishing practical deployment recommendations. All code, evaluation protocols, synthetic datasets, and analysis tools are publicly available at https://github.com/inflaton/df4tir.

## II. RELATED WORK

### A. Multi-Agent LLM Systems and Coordination

Berrich et al. [11] demonstrated that organizational approaches to multi-agent systems can improve task execution reliability. Recent advances in LLM-based multi-agent systems have demonstrated significant potential for complex task automation. Acharya et al. provide a comprehensive survey of agentic AI systems, emphasizing the importance of systematic evaluation methodologies [1]. Hong et al. introduced MetaGPT, demonstrating structured multi-agent coordination through role specialization [2], while Chawla et al. identify tool use as a fundamental pillar of agentic AI systems [9].

### B. Tool-Use Evaluation and Error Analysis

Understanding failure modes in tool-augmented LLMs represents an emerging but critical research area. Kokane et al. argue for moving beyond simple success rates to systematic error characterization, introducing SpecTool with seven error categories for single-agent tool-use evaluation [8]. Their work demonstrates that even state-of-the-art models exhibit systematic tool-use failures, but focuses on single-agent scenarios without addressing the coordination challenges inherent in multi-agent systems.

### C. Edge Deployment and Model Optimization

The deployment of LLMs on resource-constrained hardware has garnered significant attention due to privacy and cost considerations. Frantar et al. and Dettmers et al. demonstrate that 4-bit quantization can dramatically reduce memory requirements while maintaining performance for single-model inference [12], [13]. Huang and Wang extend this work to multi-model scenarios, showing that medium-sized models (7B–32B parameters) achieve optimal performance-efficiency trade-offs on edge hardware [10].

### D. LLM Reliability and Failure Analysis

Despite substantial research on LLM performance evaluation [6], [7], [14], systematic diagnostic frameworks for multi-agent reliability remain critically underdeveloped. Traditional evaluation methodologies prioritize aggregate success metrics—such as task completion rates or accuracy scores—that obscure the underlying causes of system failures [8]. While recent work has begun examining single-agent tool-use errors [8] and benchmarking capabilities on isolated tasks [4], [5], these approaches do not address the coordination challenges and cascading failure modes inherent in multi-agent architectures.

## III. METHODOLOGY

### A. Diagnostic Framework Overview

Our diagnostic framework provides systematic methodology for evaluating procedural reliability in multi-agent LLM systems through three core components: (1) structured error taxonomy for comprehensive failure characterization, (2) standardized evaluation protocols for reproducible assessment, and (3) hardware-performance characterization for deployment guidance.

The framework is designed for broad applicability across multi-agent scenarios while maintaining diagnostic precision through domain-specific instantiation. We demonstrate this approach using invoice reconciliation as a representative case study of structured business automation requiring multi-modal processing and coordinated tool use.

### B. Multi-Agent System Architecture

Building upon our previous work on edge-deployed LLM systems for invoice reconciliation [15], we implement a modular three-agent architecture orchestrated through LangGraph. While the prior work demonstrated practical feasibility and resource efficiency of edge deployment, the current work extends this foundation by introducing comprehensive diagnostic capabilities to systematically characterize procedural reliability across model scales and hardware configurations.

The architecture comprises three specialized agents, each enhanced with detailed instrumentation for failure diagnosis:

- **Email Agent:** Processes input documents and manages OCR tool invocation for multi-modal content analysis. Enhanced with logging mechanisms to capture OCR initialization failures, parameter mismatches, execution errors, and result interpretation issues.

- **Data Engineering Agent:** Executes structured database operations through DB Query Tool and DB Update Tool with validation and error handling. Instrumented to identify query construction failures, schema validation errors, and update operation anomalies.

- **Reconciliation Agent:** Coordinates workflow execution, processes agent outputs, and makes final reconciliation decisions. Monitors inter-agent communication failures and coordination breakdowns.

### C. Error Taxonomy for Multi-Agent Tool Use

Our diagnostic framework centers on a systematic error taxonomy that provides fine-grained characterization of tool-use failures in multi-agent systems. The taxonomy systematically combines four error types with three tool categories, yielding twelve distinct failure modes.

**Error Type Categories:**

- **Not Initialized:** Agent fails to properly invoke tool due to malformed calls, incorrect tool names, or invalid parameter structures.

- **Arguments Mismatch:** Tool invocation succeeds but contains incorrect or invalid arguments leading to tool-specific failures.

- **Error:** Tool executes with valid parameters but returns runtime errors or exceptions.

- **Result Mismatch:** Tool execution succeeds but produces outputs that cause downstream task failure.

**Tool Categories:** OCR processing, database query operations, and database update operations, representing common primitives in structured business automation for invoice reconciliation.

**Complete Taxonomy:** The cross-product of four error types and three tool categories produces 12 diagnostic categories: OCR_TOOL_NOT_INITIALIZED,

OCR_TOOL_ARGS_MISMATCH, OCR_TOOL_ERROR, OCR_TOOL_RESULT_MISMATCH, DB_QUERY_TOOL_NOT_INITIALIZED, DB_QUERY_TOOL_ARGS_MISMATCH, DB_QUERY_TOOL_ERROR, DB_QUERY_TOOL_RESULT_MISMATCH, DB_UPDATE_TOOL_NOT_INITIALIZED, DB_UPDATE_TOOL_ARGS_MISMATCH, DB_UPDATE_TOOL_ERROR, and DB_UPDATE_TOOL_RESULT_MISMATCH.

*D. Evaluation Dataset*

We establish standardized evaluation protocols using a carefully constructed synthetic dataset mirroring real-world invoice reconciliation scenarios. The dataset comprises 1,980 test instances balanced between vision-enabled (990) and text-only (990) tasks.

*E. LLM Integration*

**Model Selection:** To establish comprehensive reliability benchmarks across the performance spectrum, we evaluate two categories of LLMs:

• **Closed-Source Models:** OpenAI's GPT-4 series (gpt-4o, gpt-4o-mini, gpt-4.1-nano, gpt-4.1-mini, gpt-4.1) and Anthropic's Claude models (claude-3-5-sonnet, claude-3-7-sonnet) serve as reference baselines representing state-of-the-art proprietary systems.

• **Open-Weight Models:** The Qwen2.5 series (3B, 7B, 14B, 32B, 72B) and Functionary V3.1 (8B, 70B) offer systematic coverage across model scales, with local deployment ensuring data sovereignty.

**Deployment Configuration:** All open-weight models were executed locally using Ollama v0.6.8 with 4-bit quantization (Q4_K_M format) to simulate realistic edge deployment constraints while maintaining acceptable inference quality. Throughout this paper, open-weight models follow the official naming conventions from Ollama (e.g., qwen2.5:3b, qwen2.5:14b, functionary-small).

**Hardware Infrastructure:** To evaluate deployment trade-offs across diverse environments, all open-weight models are tested on three representative edge platforms: NVIDIA RTX A6000 Ada GPU (48GB VRAM, ~$10K), NVIDIA GeForce RTX 4090 Laptop GPU (16GB VRAM, ~$5K), and Apple M3 Max (96GB unified memory, ~$4K).

*F. Evaluation Metrics and Protocol*

Our evaluation employs four complementary metrics: Success Rate (SR), Execution Time, Process Steps, and OCR F1 Score. All experiments employ deterministic execution (temperature=0, fixed prompts) across 1,980 test instances.

*G. Qualitative Failure Analysis*

To complement our quantitative error taxonomy, we conduct systematic qualitative analysis of failure patterns. This analysis focuses specifically on initialization failures—the dominant error type revealed by our taxonomy—to identify the underlying reasoning deficiencies that lead to systematic tool-use breakdowns.

## IV. RESULTS AND ANALYSIS

*A. Performance and Reliability Trade-offs*

Table I summarizes overall performance metrics for all evaluated LLM configurations1.

TABLE I
OVERALL INVOICE RECONCILIATION PERFORMANCE ACROSS LLM CONFIGURATIONS

| Model | Platform | SR (%) | Time (s) | Steps |
|---|---|---|---|---|
| *Closed-source models* | | | | |
| gpt-4o-mini | OpenAI | 99.3 | **7.6 ± 1.6** | 8.6 ± 1.2 |
| gpt-4o | OpenAI | 98.8 | 12.4 ± 13.2 | 8.5 ± 0.6 |
| gpt-4.1-nano | OpenAI | 85.7 | 8.6 ± 12.3 | 18.5 ± 18.6 |
| gpt-4.1-mini | OpenAI | 99.5 | 8.6 ± 2.7 | 9.6 ± 1.9 |
| gpt-4.1 | OpenAI | **100.0** | 8.3 ± 7.7 | 9.9 ± 9.7 |
| claude-3-5-sonnet | Anthropic | 97.8 | 18.7 ± 6.3 | 8.5 ± 0.5 |
| claude-3-7-sonnet | Anthropic | 99.3 | 16.8 ± 12.3 | 8.5 ± 0.7 |
| *Open-weight models* | | | | |
| qwen2.5:3b | RTX A6000 | 13.9 | 8.1 ± 19.0 | 18.8 ± 14.7 |
| qwen2.5:7b | RTX A6000 | 57.3 | 9.9 ± 10.2 | 20.7 ± 22.0 |
| qwen2.5:14b | RTX A6000 | 96.6 | **7.3 ± 8.0** | 8.6 ± 2.2 |
| qwen2.5:32b | RTX A6000 | **100.0** | 13.9 ± 16.9 | 8.5 ± 0.5 |
| qwen2.5:72b | RTX A6000 | 95.1 | 39.3 ± 24.8 | 8.7 ± 4.2 |
| functionary-small | RTX A6000 | 7.5 | 4.0 ± 21.6 | 5.6 ± 3.7 |
| functionary-medium | RTX A6000 | 53.2 | 30.8 ± 242.6 | 7.8 ± 2.8 |

**Note:** SR = Success Rate. Execution times include full network latency for API-based models. All metrics based on 1,980 test instances.

**Reliability Thresholds:** Our analysis reveals clear performance thresholds informing deployment decisions:

• **Flawless performance:** qwen2.5:32b achieves perfect 100% success matching GPT-4.1, demonstrating that open-weight models can achieve closed-source reliability at sufficient scale.

• **Production threshold:** qwen2.5:14b maintains 96.6–97.4% success across all hardware platforms, establishing a practical reliability threshold for cost-sensitive deployments.

• **Diminishing returns:** qwen2.5:72b exhibits slightly lower success (95.1%) than its 32B counterpart, suggesting task-specific capacity limits.

• **Fundamental limits:** Smaller models exhibit significant performance degradation—qwen2.5:7b achieves 49.3–58.5%, while qwen2.5:3b drops to 13.1–14.9%.

**Hardware Impact on Deployment Viability:** Platform selection critically affects both latency and reliability:

• **Hardware optimization:** qwen2.5:14b executes in 7.3 seconds on RTX A6000 versus 60.0 seconds on M3 Max—an 8.2× difference highlighting optimized hardware importance.

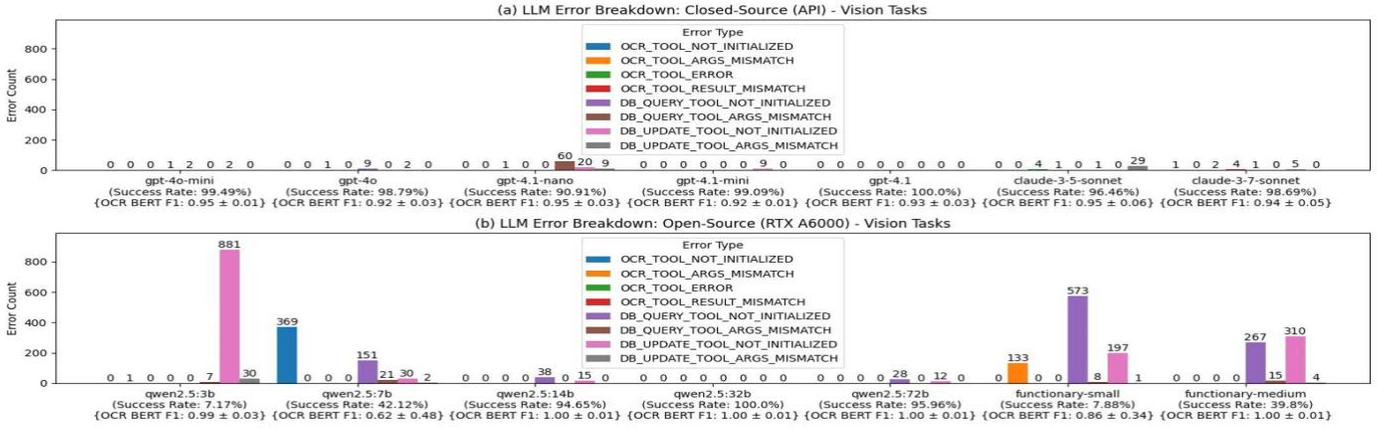

Figure 1: Systematic error characterization through our diagnostic taxonomy. (a) Closed-source models demonstrate superior procedural reliability with minimal errors. (b) Open-weight models show dramatic variation with scale; initialization failures dominate smaller models. The taxonomy reveals specific targets for system improvement beyond aggregate success metrics.

• **Planning efficiency matters:** Hardware alone is insufficient. qwen2.5:7b on M3 Max averages 85.9 seconds despite smaller size, requiring 20.7 steps per task versus 8.6 steps for the larger 14B model.

• **Pathological behavior:** Execution logs show that qwen2.5:7b often enters cyclical decision loops, occasionally exceeding LangGraph's recursion limit (25 steps) and triggering forced termination. This suggests that planning efficiency improves with model capacity.

*Tiered Deployment Recommendations:* Our framework supports evidence-based deployment across organizational constraints:

1) **Maximum Reliability:** qwen2.5:32b on RTX A6000 ($10K) achieves flawless performance comparable to GPT-4.1 while maintaining data sovereignty.

2) **Balanced Performance:** qwen2.5:14b on RTX 4090 ($5K) provides 96.6% reliability—the optimal choice for most SME deployments.

3) **Budget-Constrained:** Models below 14B require additional validation layers, automated retry mechanisms, and acceptance of higher failure rates for production use.

*Summary:* These results demonstrate that while closed-source LLMs maintain a slight edge in some configurations, the best open-weight models at the 32B scale achieve parity for invoice reconciliation tasks. For latency-sensitive applications, qwen2.5:14b on RTX A6000 offers an optimal balance of accuracy (96.6%) and speed (7.3 seconds), outperforming even the fastest API-based models. These findings are encouraging for deploying open-weight LLMs in sensitive financial workflows, especially when paired with capable hardware.

### B. Diagnostic Error Analysis

Tables II and III present detailed error distributions across all evaluated models, categorized according to our 12-category taxonomy. Figure 1 visualizes these patterns, providing unprecedented insight into multi-agent reliability.

*Systematic Failure Patterns:* Tool initialization failures (DB_UPDATE_TOOL_NOT_INITIALIZED and DB_QUERY_TOOL_NOT_INITIALIZED) dominate error distributions across all model scales. This pattern holds consistently for both vision and non-vision tasks, indicating that procedural invocation—not parameter handling or result interpretation—represents the primary reliability bottleneck.

TABLE II
ERROR COUNTS AND RATES FOR VISION TASKS ACROSS EIGHT FAILURE CATEGORIES

| Platform | Model | OCR Init | OCR Args | OCR Err | OCR Res | DB Q Init | DB Q Args | DB U Init | DB U Args |
|---|---|---|---|---|---|---|---|---|---|
| RTX A6000 | qwen2.5:3b | 0 | 1 | 0 | 0 | 0 | 7 | 881 | 30 |
| RTX A6000 | qwen2.5:7b | 369 | 0 | 0 | 0 | 151 | 21 | 30 | 2 |
| RTX A6000 | qwen2.5:14b | 0 | 0 | 0 | 0 | 38 | 0 | 15 | 0 |
| RTX A6000 | qwen2.5:32b | 0 | 0 | 0 | 0 | 0 | 0 | 0 | 0 |
| RTX A6000 | qwen2.5:72b | 0 | 0 | 0 | 0 | 28 | 0 | 12 | 0 |
| RTX A6000 | func-small | 0 | 133 | 0 | 0 | 573 | 8 | 197 | 1 |
| RTX A6000 | func-medium | 0 | 0 | 0 | 0 | 267 | 15 | 310 | 4 |
| OpenAI | gpt-4o-mini | 0 | 0 | 0 | 1 | 2 | 0 | 2 | 0 |
| OpenAI | gpt-4o | 0 | 0 | 1 | 0 | 9 | 0 | 2 | 0 |
| OpenAI | gpt-4.1-nano | 0 | 0 | 1 | 0 | 0 | 60 | 20 | 9 |
| OpenAI | gpt-4.1-mini | 0 | 0 | 0 | 0 | 0 | 0 | 9 | 0 |
| OpenAI | gpt-4.1 | 0 | 0 | 0 | 0 | 0 | 0 | 0 | 0 |
| Anthropic | claude-3-5 | 0 | 0 | 4 | 1 | 0 | 1 | 0 | 29 |
| Anthropic | claude-3-7 | 1 | 0 | 2 | 4 | 1 | 0 | 5 | 0 |

Note: OCR Init = OCR_TOOL_NOT_INITIALIZED, OCR Args = OCR_TOOL_ARGS_MISMATCH, OCR Err = OCR_TOOL_ERROR, OCR Res = OCR_TOOL_RESULT_MISMATCH, DB Q Init = DB_QUERY_TOOL_NOT_INITIALIZED, DB Q Args = DB_QUERY_TOOL_ARGS_MISMATCH, DB U Init = DB_UPDATE_TOOL_NOT_INITIALIZED, DB U Args = DB_UPDATE_TOOL_ARGS_MISMATCH. Percentages based on 990 vision tasks.

*Model Scale Impact:* Error rates demonstrate clear inverse correlation with model parameters among open-weight models. qwen2.5:3b exhibits 881 database update initialization errors (88.99% of tasks); qwen2.5:14b reduces this to 15–38 errors (1.5–3.8%); qwen2.5:32b achieves zero errors across all categories, matching GPT-4.1.

TABLE III
ERROR COUNTS AND RATES FOR NON-VISION TASKS ACROSS FOUR DATABASE FAILURE CATEGORIES

| Platform | Model | DB Q Init | DB Q Args | DB U Init | DB U Args |
|---|---|---|---|---|---|
| RTX A6000 | qwen2.5:3b | 7 (0.71%) | 17 (1.72%) | 756 (76.36%) | 5 (0.51%) |
| RTX A6000 | qwen2.5:7b | 264 (26.67%) | 0 (0.00%) | 7 (0.71%) | 1 (0.10%) |
| RTX A6000 | qwen2.5:14b | 8 (0.81%) | 1 (0.10%) | 4 (0.40%) | 1 (0.10%) |
| RTX A6000 | qwen2.5:32b | 0 (0.00%) | 0 (0.00%) | 0 (0.00%) | 0 (0.00%) |
| RTX A6000 | qwen2.5:72b | 57 (5.76%) | 0 (0.00%) | 0 (0.00%) | 0 (0.00%) |
| RTX A6000 | functionary-small | 806 (81.41%) | 4 (0.40%) | 108 (10.91%) | 2 (0.20%) |
| RTX A6000 | functionary-medium | 187 (18.89%) | 12 (1.21%) | 131 (13.23%) | 1 (0.10%) |
| OpenAI | gpt-4o-mini | 9 (0.91%) | 0 (0.00%) | 0 (0.00%) | 0 (0.00%) |
| OpenAI | gpt-4o | 0 (0.00%) | 0 (0.00%) | 11 (1.11%) | 0 (0.00%) |
| OpenAI | gpt-4.1-nano | 55 (5.56%) | 24 (2.42%) | 114 (11.52%) | 1 (0.10%) |
| OpenAI | gpt-4.1-mini | 0 (0.00%) | 0 (0.00%) | 0 (0.00%) | 0 (0.00%) |
| OpenAI | gpt-4.1 | 0 (0.00%) | 0 (0.00%) | 0 (0.00%) | 0 (0.00%) |
| Anthropic | claude-3-5-sonnet | 0 (0.00%) | 0 (0.00%) | 0 (0.00%) | 9 (0.91%) |
| Anthropic | claude-3-7-sonnet | 0 (0.00%) | 0 (0.00%) | 1 (0.10%) | 0 (0.00%) |

**Note:** DB Q Init = DB_QUERY_TOOL_NOT_INITIALIZED, DB Q Args = DB_QUERY_TOOL_ARGS_MISMATCH, DB U Init = DB_UPDATE_TOOL_NOT_INITIALIZED, DB U Args = DB_UPDATE_TOOL_ARGS_MISMATCH. Percentages based on 990 non-vision tasks.

### C. Qualitative Error Analysis

To complement our quantitative taxonomy, we conducted systematic qualitative analysis of initialization failures—the dominant error type across smaller models. We manually inspected 200 randomly sampled failure cases from execution logs of qwen2.5:3b, qwen2.5:7b, and Functionary-Small to identify underlying reasoning patterns.

*Failure Mode Classification:* Our qualitative review identifies two dominant failure patterns:

**Omission Failures (~68% of sampled cases).** Agents fail to recognize that tool invocation is required, producing natural-language responses instead. For instance, when prompted to execute a database query, qwen2.5:7b occasionally generates:

> "Based on the invoice information provided, the payment amount is $1,234.56 for invoice ID INV-2024-001..."

rather than invoking the required db_query_tool function. This reflects inadequate procedural reasoning—the model conceptually resolves the task but omits the necessary operational step, revealing weak coupling between semantic understanding and action execution.

**Malformed Call Failures (~32% of sampled cases).** Models correctly identify tool-use needs but construct invalid function calls through:

• Incorrect tool names (e.g., database_query instead of db_query_tool)

• Invalid JSON structure (missing parameters or malformed syntax)

• Hallucinated parameters (non-existent fields or incorrect data types)

These patterns indicate insufficient schema grounding and weak syntactic discipline during tool invocation.

*Model-Specific Error Distributions:* Failure patterns vary systematically across architectures:

• **qwen2.5:3b:** Dominated by omission failures (~89% of sampled errors), reflecting shallow procedural reasoning and weak context maintenance.

• **qwen2.5:7b:** Predominantly omission failures with high argument precision when tools are invoked.

• **Functionary-Small:** Balanced distribution (~55% omission, 45% malformed).

Overall, our qualitative analysis reveals that tool-calling failures in smaller models stem from two main issues: failure to initiate tool use and incorrect execution of tool calls. While larger models like qwen2.5:7b show improved precision, they still suffer from procedural lapses such as omission errors. Reliable multi-agent systems will require not just capable models but also robust orchestration mechanisms, schema-aware invocation, and strategies for error recovery to mitigate these recurring failure modes.

## V. CONCLUSION

This work introduces a systematic diagnostic infrastructure for evaluating tool-use reliability in multi-agent LLM systems, addressing a critical gap in existing evaluation frameworks. Our 12-category error taxonomy enables fine-grained characterization of procedural failures, revealing actionable insights often obscured by aggregate success metrics.

An analysis of 1,980 deterministic test instances yields three key findings: (1) Tool initialization failures—rather than parameter tuning or output interpretation—emerge as the primary reliability bottleneck, occurring at catastrophic rates in small models (e.g., qwen2.5:3b: 89% errors) and entirely absent in large models (e.g., qwen2.5:32b, GPT-4.1: 0% errors); (2) Distinct capacity thresholds are observed at 14B parameters (minimum viable, with 96.6–97.4% success) and 32B parameters (achieving reliability parity with closed-source models); (3) Hardware exerts a substantial influence, with up to 8.2× latency variation across platforms.

Based on our findings, we propose tiered deployment recommendations: qwen2.5:32b on RTX A6000 ($10K) delivers flawless performance matching GPT-4.1; qwen2.5:14b on RTX 4090 ($5K) offers strong accuracy-efficiency trade-offs suitable for most applications; sub-14B models require substantial architectural augmentation for acceptable reliability.

Future directions include: (1) adversarial and negative-case protocol design; (2) distributed failure analysis; (3) self-healing agent architectures informed by our taxonomy; and (4) cross-domain validation beyond finance.

As multi-agent LLM systems move from prototypes to production, ad hoc evaluation no longer suffices. Our diagnostic framework shifts evaluation from whether a system works to why it fails, providing a reproducible, fine-grained method for measuring procedural reliability and enabling more robust, predictable, and trustworthy agentic AI for real-world deployment.